\documentclass[conference]{IEEEtran}

\usepackage{cite}
\usepackage{amsmath,amssymb}
\usepackage{graphicx}
\usepackage{booktabs}
\usepackage{url}
\usepackage{xspace}
\usepackage{fontawesome5}
\usepackage[hidelinks]{hyperref}

\graphicspath{{figures/}}

\title{DeFiFlowBench: Benchmarking and Improving Safe Executability in\\
Natural-Language DeFi Workflow Synthesis}

\author{
\IEEEauthorblockN{Abhinav Rajeev Kumar,
Harshit Arora\textsuperscript{*},
Varun Singh\textsuperscript{*},
Manikandan Nanjappan\textsuperscript{\ensuremath{\dagger}}}
\IEEEauthorblockA{SRM Institute of Science and Technology\\
\{ak0929, ha8052, vs6320, manikann6\}@srmist.edu.in}
}

\begin{document}
\bstctlcite{koan:bibcontrol}
\maketitle
\begingroup
\renewcommand{\thefootnote}{\fnsymbol{footnote}}
\footnotetext[1]{Equal contribution.}
\footnotetext[2]{Advisor.}
\endgroup

\begin{abstract}
A structurally valid DeFi workflow can still authorize a costly trade.
We introduce \textsc{DeFiFlowBench}, a benchmark of 207 team-authored
prompts for natural-language DeFi workflow synthesis. It measures graph
coverage, configuration completeness, and declared safety predicates,
then tests supported trade configurations on a local EVM. Direct,
constrained, and few-shot prompting produce $14$--$19$ unsafe held-out
executions per configuration under a fixed $5\%$ price-impact cap.
A slippage bound derived from a quote does not prevent the price impact
of the order itself. We propose Koan-Safe, which combines a prompt-only
intent parser, a replaceable generator, and structural repair with
default safety parameters. On 75 held-out workflow prompts, its hybrid
variant scores $0.67$ on the static safety proxy, compared with $0.33$
for the best baseline. Koan-Safe records no unsafe executions on the
saved benchmark outputs. A matched-candidate ablation produces
$14$--$17$ unsafe executions when enforcement is disabled. Additional
tests expose the limits of default injection: permissive existing
thresholds can still authorize unsafe trades. A separately evaluated
policy cap addresses this failure on a 36-case diagnostic grid.
These results support explicit trade protections and execution-based
evaluation, while distinguishing declared safety from a general guarantee.

\end{abstract}


\section{Introduction}
\label{sec:intro}

Natural-language DeFi requests require a system to choose protocol
operations and bind trade parameters. They also raise a separate safety
question. A minimum output computed from a quote can enforce a slippage
tolerance while allowing a large loss from the order's own price impact.
The graph can contain the expected calls and still authorize a costly trade.

Research on natural-language workflow
generation~\cite{worfbench2025,chat2workflow2026,flowmind2023}
examines structural correctness and executable workflows.
Execution-aware transaction evaluation also tests whether generated
calls achieve intended state changes~\cite{intent2tx2026}.
We study a narrower question: do generated DeFi workflows include the
protections needed to control trade risk, and what happens when those
protections are missing?

Separating a generator from an external checker is well established in
planning and agent safety~\cite{llmmodulo,guardagent}. The question here
is domain-specific: how much can fixed workflow repairs improve declared
DeFi protections, and which failures remain in numeric policy settings?

We introduce \textsc{DeFiFlowBench}, with 120 development prompts and
87 held-out prompts covering swaps, limit orders, cross-chain transfers,
and compositional tasks. Three nested static scores measure required
graph coverage, configuration completeness, and declared safety
predicates. A separate execution check interprets supported trade
configurations against a local constant-product AMM. It does not execute
the complete generated DAG. We use \emph{safe-executability} for the
static proxy throughout; passing this score does not guarantee that a
workflow fulfills the user's intent or is safe on a deployed protocol.

On the 75 held-out workflow prompts, the best baseline scores $0.33$
on this proxy. Templates omit concrete trade parameters, while direct,
constrained, and few-shot LLM prompting often omits price-impact gates.
Each of these LLM configurations produces $14$--$19$ unsafe
high-impact executions in the local checker. These observations
motivate Koan-Safe, which separates prompt-only intent parsing,
candidate generation, and the addition of structural requirements and
default safety parameters.

Koan-Safe's hybrid variant reaches $0.67$ on the static proxy. The
rule-based and LLM variants reach $0.43$ and $0.64$. The method was
frozen before the team authored the held-out split, which includes
structures outside its presets. An enforcement-on/off comparison and a
34-pair metamorphic suite examine where the improvement comes from and
how it responds to rewritten requests. The on/off comparison reconstructs
both conditions from the same saved candidate, controlling model-output
variation without additional API calls.

The contribution is a benchmark and an empirical evaluation of a
workflow-repair method under explicit, limited execution semantics.
The corrected execution checker applies a fixed impact cap regardless
of gate declaration. A separate stress grid tests existing permissive
thresholds and a post-hoc policy-cap extension. Together, these experiments
show where structural repair helps and where it must be supplemented by
numeric policy validation. Neither a high static score nor zero observed
unsafe trades establishes a general safety guarantee.

\begin{center}
\href{https://github.com/Varun-2538/Koan}{\faGithub\enspace
\texttt{github.com/Varun-2538/Koan}}
\end{center}

\section{Problem statement and the safety ladder}
\label{sec:method}

\subsection{Task}

DeFi workflow synthesis maps a natural-language intent $p$ to a workflow
specification $W = (N, E, C)$: a set $N$ of typed nodes drawn from a fixed
executable DeFi node vocabulary (e.g.\ \texttt{tokenSelector},
\texttt{oneInchQuote}, \texttt{priceImpact\-Calculator},
\texttt{oneInchSwap}, \texttt{limitOrder}, \texttt{fusionPlus},
\texttt{transactionMonitor}), a set $E \subseteq N \times N$ of type-level
data-flow edges, and an execution configuration $C$ mapping keys such as
\texttt{fromToken}, \texttt{amount}, \texttt{slippage}, and
\texttt{targetPrice} to concrete values.

Each benchmark item pairs a prompt with an team-authored gold annotation
$(N^{\star}, E^{\star}, C^{\star}, S^{\star})$, where $S^{\star}$ is a set
of \emph{safety predicates} the workflow must satisfy to be safely
executable: a slippage bound, a price-impact gate, transaction monitoring,
a distinct bridge destination. The annotation also marks underspecified
prompts, for which the correct response is a clarification request rather
than a workflow.

\subsection{The safety ladder}
\label{sec:ladder}

We use three nested static scores and a separate execution check. The
execution check has a different denominator and can run a trade
configuration even when the full graph fails the static checks.

\paragraph{Level 1: graph-valid.} The workflow recalls every required node
type and required type-level edge, and the generator did not error. This
level is deliberately lenient. Extra nodes and edges do not affect this
score, and it does not check acyclicity or dependency semantics.

\paragraph{Level 2: executable.} The workflow is graph-valid \emph{and}
every required configuration key in $C^{\star}$ holds a concrete,
non-placeholder value. This is a configuration-completeness proxy;
it does not verify that values match the user's intent or satisfy
numeric bounds. A template with no required trade amount fails it.

\paragraph{Level 3: statically safe.} The workflow is executable
\emph{and} every required safety predicate in $S^{\star}$ is declared and
parameterized. Predicates are derived uniformly from the normalized
workflow for all systems, so no system benefits from self-reported safety.

\paragraph{Execution check.} We interpret the normalized configuration
and safety declarations in a separate local-EVM test program. This
program does not traverse the generated DAG or invoke Koan's executors.
We deploy a constant-product AMM (Uniswap-V2-style $x \cdot y = k$ with a
$0.3\%$ fee) and ERC20 tokens on an in-process EVM, seed deterministic
reserves, and submit supported swaps that pass any declared gate as local transactions. The
harness observes the mined receipt, gas, or revert and labels the outcome
(\texttt{executed\_safe}, \texttt{reverted\_slippage},
\texttt{aborted\_price\_impact}, \texttt{unsafe\_executed},
\texttt{pending\_not\_filled}, or \texttt{not\_executable}). Execution
semantics include \texttt{transferFrom} calls and \texttt{require}
checks. Tokens, reserves, and routing are simplified. Price-impact gates
run in Python before submission. We compare AMM arithmetic against
mainnet reserves in Section~\ref{sec:fidelity}.

The execution safe-rate includes blocked trades, slippage failures, and
unfilled limit orders. It excludes indeterminate and unsupported cases,
so it is not a task-success rate. Evaluator v2 marks a mined swap unsafe
whenever impact exceeds $5\%$, regardless of whether a gate is declared.
Threshold values use percentage points. A sensitivity analysis tests the
alternative interpretation of ambiguous historical numeric outputs.
Invalid explicit parameters produce a non-executable result rather than
silently disabling their protection.

A separate \emph{clarification axis} scores underspecified prompts, where
emitting a confidently wrong workflow is the failure mode.

\paragraph{Slippage is not price impact.} The distinction that drives our
headline result is between two protections that generators routinely
conflate. A \emph{slippage bound} sets a minimum output computed from the
already impact-adjusted quote; it protects against price movement between
quote and execution. It does nothing to stop a large order from executing
at severe \emph{self-inflicted} price impact. A separate
\emph{price-impact gate} can block that trade before submission if its
threshold is sufficiently restrictive. A swap above the fixed impact cap
is labeled \texttt{unsafe\_executed}, including when a permissive gate
allows it.

\section{Benchmark}
\label{sec:benchmark}

\subsection{Categories and prompts}

\textsc{DeFiFlowBench} contains 120 team-authored development prompts
across four categories: token swaps (40), limit orders (30), cross-chain
transfers (30), and compositional workflows that combine primitives (20).
Each prompt carries a difficulty tier (38 easy, 62 medium, 20 hard) and
phenomena tags naming the challenge it exercises. Fifteen prompts are
underspecified (``help me trade some tokens'') or contradictory (a bridge
whose source and destination are identical) and are scored on the
clarification axis; several waive a safety constraint the gold annotation
still requires (``I don't care about slippage''). Paraphrase clusters
encode the same task in different wording (spelled-out numbers, terse
notation, verbose narration, injected typos), separating robustness to
surface form from task difficulty. Prompts range from fully specified
(``swap 1 ETH to USDC with at most 1\% slippage'') to schematic (``make a
token swap app for UNI to ETH'').

\paragraph{Held-out test split.} To measure generalization rather than
in-distribution fit, we authored a separate 87-prompt split after freezing
the Koan-Safe method, and generated one saved output per system and prompt. It is
harder in two ways. Its canonical-structure prompts use fresh surface
forms and out-of-vocabulary token symbols (MKR, SUSHI, GRT) absent from
Koan-Safe's token dictionary. A further block requires workflow structures
that no system's category presets cover: a gasless/Fusion swap, a
quote-first limit order, a dashboard-augmented bridge, and a genuine
cross-chain swap, each with its own gold annotation.

\paragraph{Metamorphic safety suite.} Following the testing principle of
checking relations between transformed inputs~\cite{metamorphic},
a third split of 34 base/variant
prompt pairs tests whether safety survives risk-relevant rewrites of the
request. Each pair is checked against the relation its transform must
satisfy: amount monotonicity (a $100\times$ larger trade must not weaken
safety), threshold tightening, waiver resistance (an appended instruction
to skip safety checks), paraphrase invariance, and drop-field
non-fabrication (a removed critical field must not be invented). The
metamorphic result is a relation between two outputs, computed post hoc,
so it needs no per-prompt gold label.

\subsection{Gold annotations}

Each prompt is paired with a gold annotation recording (i) the required
node types, (ii) the required type-level edges, (iii) the required
execution-configuration keys, (iv) the safety predicates that must hold,
and (v) allowed extra nodes that a richer-but-correct workflow may include
without penalty. The annotation scheme is documented in the artifact's
annotation guide so the benchmark can be extended consistently. The node
vocabulary is taken from an existing executable DeFi node catalog, so
every required node corresponds to a real executor rather than an
abstract label.

\subsection{Systems under test}

All systems are scored by the same evaluator. Koan-Safe and LLM
generators read the raw prompt text. The template reference receives
the annotated category, and the oracle additionally reads gold structure.

\begin{itemize}
  \item \textbf{Oracle (ceiling).} Reconstructs a correct workflow from
        the gold annotation and fills configuration with generic values.
        Its $1.00$ static scores check scorer consistency, not semantic
        correctness of the resulting trades.
  \item \textbf{Null and Random (floor).} An empty workflow and a
        random-size sample of real nodes in a random chain with no
        configuration. Both score $0.00$, guarding against a gameable
        metric.
  \item \textbf{Template-only.} Emits the canonical workflow for the
        prompt's category with static safety defaults, but performs no
        language understanding and cannot populate trade-specific
        configuration.
  \item \textbf{Koan.} The regex-fallback intent parser and workflow
        generator of an existing deployed no-code DeFi platform, run
        offline and deterministically.
  \item \textbf{Direct / Constrained / Few-shot / Safety-instructed
        LLM.} A language model prompted to emit a workflow graph and
        configuration as JSON. The variants differ only in the prompt:
        no constraints; generic task-specific structural and safety
        constraints; a single worked example; and an explicit instruction
        to make the workflow safe to execute.
  \item \textbf{Koan-Safe (proposed).} Our enforcement system, evaluated
        with all three generators and with the enforcement layer toggled
        off as an ablation.
\end{itemize}

LLM systems run on two model families to test whether findings are
model-specific. Model outputs are normalized before scoring (node and
edge vocabularies filtered; index-based and name-based edge encodings
both accepted), and each raw response is retained so metrics can be
re-derived without further queries.

\section{Koan-Safe: structural repair and safety defaults}
\label{sec:koansafe}

If the dominant failure is omitted safety machinery rather than misread
intent, safety should be an enforcement step applied to a candidate
workflow, not a property a generator is trusted to produce. Koan-Safe
factors synthesis into three replaceable stages: an intent parser, a
candidate generator, and a generator-agnostic safety-enforcement layer.
The factoring lets the enforcement layer be toggled and measured in
isolation.

A note on naming: \emph{Koan} is the existing deployed platform we
evaluate as a baseline; \emph{Koan-Safe} is the method proposed here;
\textsc{DeFiFlowBench} is the benchmark.

\paragraph{Intent parser.} A deterministic, prompt-only parser maps the
request text to a typed intent: task category, the trade-intent fields it
can extract (tokens, amount, slippage, target price, chains), and a
decision to build or ask for clarification. It reads only the prompt
string, never the gold annotation or category label, so it has exactly
the information available to the LLM baselines. Clarification fires only
when acting would require guessing trade intent: no recognizable token,
identical source and destination, or a bridge with no destination.

\paragraph{Candidate generator.} Three interchangeable generators feed
the same enforcement layer. \emph{Rules} emits a category-appropriate
node graph and fills configuration from the parsed intent. \emph{LLM}
uses the same neural backend as the direct baseline. \emph{Hybrid} takes
the LLM's structure but backfills trade-intent configuration that the
parser read from the prompt and the model dropped.

\paragraph{Safety-enforcement layer.} Given any candidate workflow and
the parsed intent, the layer repairs structure by adding the category's
required safety nodes and reconnecting the required spine, then injects
conservative safety policy the request did not pin down: a default
slippage bound, a price-impact gate with a concrete threshold, a bridge
confirmation count, a default order expiry, and a self-recipient for
bridges. The layer does not add trade-intent fields itself. It preserves
the candidate's configuration, including any unsupported values proposed
by the LLM. The hybrid backfills missing fields from the parser but does
not verify every existing value against the request. Safety defaults are
added only when keys are absent; existing permissive thresholds are not
clamped. Waiver resistance therefore depends on the candidate and cannot
be guaranteed by this implementation.

\paragraph{Integrity of the comparison.} Koan-Safe encodes engineered
domain knowledge (a token dictionary, per-category node presets,
conservative safety defaults) but has no access to gold annotations.
Because its structural presets are fixed, the held-out split deliberately
requires workflow structures outside those presets, so generalization is
tested rather than assumed.

\section{Experimental setup}
\label{sec:setup}

We ask whether graph coverage predicts executable trade protection,
whether Koan-Safe improves the static safety proxy on new workflow
structures, and whether enforcement changes outcomes when the candidate
is held fixed.

\paragraph{Splits and models.} The development split contains 120 prompts,
including 105 workflow prompts. The held-out split contains 87 prompts,
including 75 workflow prompts, and was authored after the method was
frozen. Unless stated otherwise, results use the held-out split.
The original generation runs used temperature zero on Gemini~3.1~Flash~Lite
and GPT-5.4~mini through OpenRouter. The present evaluation reuses those
saved outputs without resampling. Static rates include a 95\% Wilson
interval in Table~\ref{tab:main-results}. These intervals summarize prompt
variation, not variation across repeated model calls.

\paragraph{Corrected execution protocol.} All reported execution results
use evaluator v2. It applies a fixed $5\%$ impact cap independently of
gate declarations, interprets thresholds in percentage points, handles
explicit percent strings, and rejects non-finite or invalid parameters.
Transactions use an explicit gas limit so a failed submitted call has a
mined receipt. Historical generation prompts did not fix numeric units;
we therefore also replay ambiguous bare thresholds under the fractional
interpretation. The artifact preserves both evaluation versions.
Equivalent configurations reuse a local result after checking the same
category, status, configuration, and safety declarations.

\paragraph{Calibration.} The oracle supplies required structure and
generic configuration values, giving $1.00$ on the static levels. This
checks scorer consistency, not fulfillment of the prompt's intended
trade. The null and random-node baselines score $0.00$ on these levels.

\section{Results}
\label{sec:results}

\begin{table*}[t]
\centering
\caption{\textbf{Main results on the held-out test split} (${n{=}75}$ workflow prompts). Static levels of the safety ladder (graph-valid, executable, declared safety proxy with a 95\% Wilson interval) and on-chain outcomes on the local EVM: Unsafe counts workflows that mined at $>$5\% own-trade price impact regardless of gate declaration; safe-rate is over prompts with a definite outcome under evaluator v2, including refusals and unfilled orders. Oracle and null/random baselines calibrate the ceiling and floor. LLM systems run at temperature~0; the best non-reference safe rate is \textbf{bold}.}
\label{tab:main-results}
\begin{tabular}{lccccc}
\toprule
 & \multicolumn{3}{c}{Static safety-ladder levels} & \multicolumn{2}{c}{On-chain (local EVM)} \\
\cmidrule(lr){2-4}\cmidrule(lr){5-6}
System & Graph-valid & Executable & Safety proxy [CI] & Unsafe & Safe-rate \\
\midrule
Oracle (ceiling) & 1.00 & 1.00 & 1.00 [0.95,1.00] & 0 & 1.00 \\
Null (floor) & 0.00 & 0.00 & 0.00 [0.00,0.05] & -- & -- \\
Random nodes & 0.00 & 0.00 & 0.00 [0.00,0.05] & -- & -- \\
\midrule
Template-only & 0.52 & 0.00 & 0.00 [0.00,0.05] & -- & -- \\
Koan (regex$+$gen) & 0.05 & 0.00 & 0.00 [0.00,0.05] & -- & -- \\
Direct LLM (Gemini 3.1 FL) & 0.09 & 0.05 & 0.03 [0.01,0.09] & 19 & 0.67 \\
Direct LLM (GPT-5.4 mini) & 0.01 & 0.00 & 0.00 [0.00,0.05] & 14 & 0.65 \\
Constrained LLM (Gemini 3.1 FL) & 0.31 & 0.25 & 0.13 [0.07,0.23] & 18 & 0.67 \\
Constrained LLM (GPT-5.4 mini) & 0.15 & 0.12 & 0.01 [0.00,0.07] & 17 & 0.63 \\
Few-shot LLM (Gemini 3.1 FL) & 0.17 & 0.15 & 0.12 [0.06,0.21] & 19 & 0.66 \\
Few-shot LLM (GPT-5.4 mini) & 0.19 & 0.13 & 0.11 [0.06,0.20] & 17 & 0.70 \\
Safety-instruct LLM (Gemini 3.1 FL) & 0.33 & 0.33 & 0.33 [0.24,0.45] & 0 & 1.00 \\
Safety-instruct LLM (GPT-5.4 mini) & 0.07 & 0.07 & 0.07 [0.03,0.15] & 0 & 1.00 \\
\midrule
Koan-Safe (rules) & 0.52 & 0.43 & 0.43 [0.32,0.54] & 0 & 1.00 \\
Koan-Safe (LLM) (Gemini 3.1 FL) & 0.71 & 0.64 & 0.64 [0.53,0.74] & 0 & 1.00 \\
Koan-Safe (LLM) (GPT-5.4 mini) & 0.61 & 0.48 & 0.48 [0.37,0.59] & 0 & 1.00 \\
Koan-Safe (hybrid) (Gemini 3.1 FL) & 0.73 & 0.67 & \textbf{0.67} [0.55,0.76] & 0 & 1.00 \\
Koan-Safe (hybrid) (GPT-5.4 mini) & 0.64 & 0.57 & 0.57 [0.46,0.68] & 0 & 1.00 \\
\bottomrule
\end{tabular}
\end{table*}

\begin{figure*}[t]
\centering
\includegraphics[width=\textwidth]{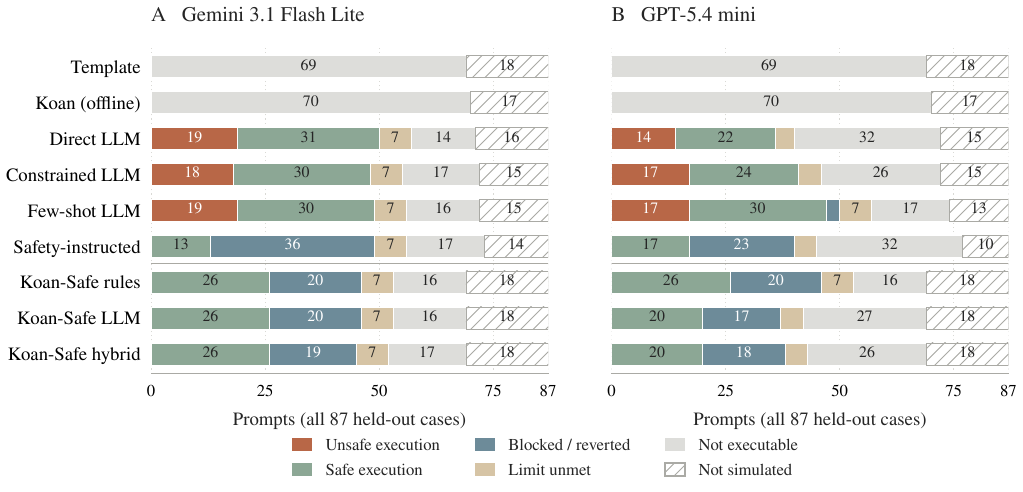}
\caption{Local execution outcomes for both model families under evaluator
v2. Every bar accounts for all 87 held-out prompts; hatched segments retain
cases the checker does not simulate. Deterministic baselines are repeated
for comparison. Counts are printed for segments of at least seven prompts.
Completed swaps and limit fills satisfy the checker's impact or
target-price condition, respectively, not the full workflow's intent.
Aborts, reverts, and unfilled limit conditions are separate from
successful execution.}
\label{fig:fates}
\end{figure*}

\begin{figure*}[t]
\centering
\includegraphics[width=\textwidth]{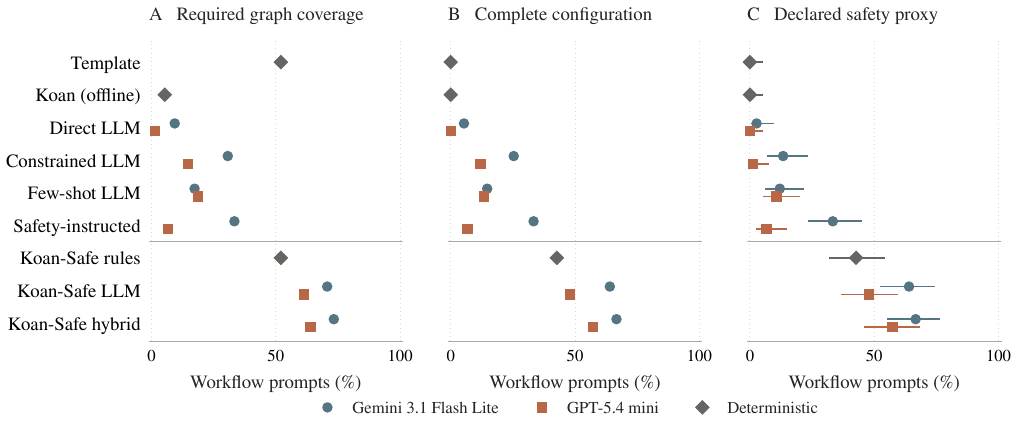}
\caption{Three nested static scores on the same 75 held-out workflow
prompts. Model families share axes; deterministic systems appear once.
Panel C includes 95\% Wilson intervals over prompts, not uncertainty
across repeated generations. The final score requires declared safety
predicates and complete configuration; it does not certify execution
or semantic fidelity.}
\label{fig:ladder}
\end{figure*}

\subsection{RQ1: Structure and execution measure different failures}
\label{sec:rq1}

No baseline exceeds $0.33$ on the static safety proxy. The
safety-instructed Gemini baseline reaches this rate with a 95\% interval
of $[0.24,0.45]$. Template-only generation covers the required graph on
$0.52$ of prompts but supplies no executable trade configuration.
The offline Koan baseline covers the graph on $0.05$ and also has zero
configuration-complete workflows. Their lack of unsafe executions
therefore reflects lack of execution.

Direct, constrained, and few-shot LLM configurations instead produce
$14$--$19$ unsafe held-out executions each. These swaps exceed the
checker's fixed impact cap. The slippage minimum is computed from the
already impact-adjusted quote, so meeting that minimum does not protect
against the order's own impact. Explicit safety instructions remove
observed unsafe executions on the held-out split, but leave static
coverage at $0.33$ for Gemini and $0.07$ for GPT.

\subsection{RQ2: Koan-Safe improves the static safety proxy}
\label{sec:rq2}

Koan-Safe hybrid with Gemini reaches $50/75$, or $0.67$, on the static
proxy, with a 95\% interval of $[0.55,0.76]$. Rules reaches $0.43$ and
the LLM variant $0.64$. Every frozen Koan-Safe configuration records
zero unsafe executions in the corrected replay. The hybrid performs
26 successful local executions, aborts 19 trades on price impact, and
leaves seven limit conditions unfilled. Its remaining outputs have
invalid or missing execution inputs or unsupported categories.

The frozen rules presets lose structural coverage on the new variants,
falling from $1.00$ on development to $0.52$ on held-out prompts.
LLM and hybrid reach $0.71$ and $0.73$ on held-out graph coverage.
Koan-Safe correctly requests clarification on nine of twelve relevant
prompts; the three misses remain unchanged.

An exploratory exact paired McNemar test compares the hybrid with
safety-instructed Gemini on the static proxy. The hybrid alone succeeds
on 26 prompts, and the baseline alone on one, giving unadjusted
$p=4.17\times10^{-7}$. Against Koan-Safe LLM, however, the hybrid adds
only two successes and loses none, giving $p=0.5$. We do not claim that
the hybrid reliably outperforms the LLM variant.

\subsection{RQ3: Matched-candidate enforcement ablation}
\label{sec:rq3}

For each Koan-Safe configuration, we reconstruct the candidate from the
saved enforcement-on response. Reapplying the frozen layer must
reproduce the saved workflow exactly. We then score and execute that
same candidate with enforcement enabled and disabled. This comparison
requires no new model calls and avoids the model variation present in
the earlier independently sampled ablation.

For Gemini, enforcement raises the LLM static proxy from $2/75$ to
$48/75$ and the hybrid from $1/75$ to $50/75$. Across both model families
and the rules generator, disabling the layer produces $14$--$17$ unsafe
held-out executions; enabling it produces zero
in these matched comparisons (Table~\ref{tab:enforcement-ablation}).
This isolates the layer's effect under the stated execution semantics,
not the correctness of full deployed workflows.

\begin{table*}[t]
\centering
\caption{Matched-candidate enforcement ablation under evaluator v2. Both conditions use the same saved candidate and parser. Static scores use workflow prompts; unsafe counts use supported local executions.}
\label{tab:enforcement-ablation}
\begin{tabular}{lcccc}
\toprule
 & \multicolumn{2}{c}{Static safety proxy} & \multicolumn{2}{c}{Unsafe executions} \\
System & off & on & off & on \\
\midrule
Koan-Safe (rules) & 0.12 & 0.43 & 17 & 0 \\
Koan-Safe (LLM) (Gemini 3.1 FL) & 0.03 & 0.64 & 17 & 0 \\
Koan-Safe (LLM) (GPT-5.4 mini) & 0.00 & 0.48 & 14 & 0 \\
Koan-Safe (hybrid) (Gemini 3.1 FL) & 0.01 & 0.67 & 16 & 0 \\
Koan-Safe (hybrid) (GPT-5.4 mini) & 0.00 & 0.57 & 16 & 0 \\
\bottomrule
\end{tabular}
\end{table*}

\subsection{Metamorphic robustness}
\label{sec:metamorphic}

The 34-pair suite evaluates amount increases, threshold tightening,
safety waivers, paraphrases, and removal of required fields.
Table~\ref{tab:metamorphic} uses evaluator-v2 units consistently for both
execution and relation checking. Direct LLMs violate 15 of 32 applicable
pairs for Gemini and 16 of 33 for GPT. Safety instructions reduce these
to four of 30 and two of 28.

The rules generator, GPT LLM variant, and GPT hybrid satisfy all
34 relations. Gemini LLM violates two threshold relations; Gemini hybrid
violates one paraphrase relation. These failures are retained. In
particular, zero unsafe executions does not imply that a system respects
every requested change in safety policy.

\begin{table*}[t]
\centering
\caption{Metamorphic safety violations (count of violated pairs, lower is better) by relation on the metamorphic suite. Amount: $100\times$ larger trade must not weaken safety; Thr: tightening tolerance must keep the gate; Waiv: an explicit safety waiver must not remove mandatory gates; Para: paraphrase must preserve class and policy; Drop: a removed field must not be fabricated. Denominators vary because pairs whose base prompt a system fails to build are not applicable.}
\label{tab:metamorphic}
\begin{tabular}{lcccccc}
\toprule
System & Amt & Thr & Waiv & Para & Drop & All \\
\midrule
Direct LLM (GPT-5.4 mini) & 7 & 0 & 7 & 2 & 0 & 16/33 \\
Direct LLM (Gemini 3.1 FL) & 8 & 1 & 6 & 0 & 0 & 15/32 \\
Safety-instruct LLM (GPT-5.4 mini) & 0 & 1 & 0 & 1 & 0 & 2/28 \\
Safety-instruct LLM (Gemini 3.1 FL) & 0 & 1 & 1 & 2 & 0 & 4/30 \\
Koan-Safe (rules) & 0 & 0 & 0 & 0 & 0 & 0/34 \\
Koan-Safe (LLM) (GPT-5.4 mini) & 0 & 0 & 0 & 0 & 0 & 0/34 \\
Koan-Safe (LLM) (Gemini 3.1 FL) & 0 & 2 & 0 & 0 & 0 & 2/34 \\
Koan-Safe (hybrid) (GPT-5.4 mini) & 0 & 0 & 0 & 0 & 0 & 0/34 \\
Koan-Safe (hybrid) (Gemini 3.1 FL) & 0 & 0 & 0 & 1 & 0 & 1/34 \\
\bottomrule
\end{tabular}
\end{table*}

\subsection{Policy stress test and post-hoc repair}
\label{sec:stress}

Default injection preserves existing parameter values. We test this
boundary on a full grid of four trade amounts and nine threshold inputs,
including missing, fractional-percent, permissive, and invalid values.
The frozen layer produces two unsafe executions, ten safe executions,
twelve protective aborts, and twelve non-executable outputs among
36 cases.

A separate policy extension caps impact thresholds at $3\%$, preserves
stricter valid bounds, replaces invalid thresholds, and caps slippage at
$1\%$. It changes no amount, token, price, or chain field. On the same
grid it produces 16 safe executions and 20 protective aborts, with no
unsafe or non-executable outcomes. We also replay this extension on the
saved model candidates. Because it was designed after the audit, this
is an engineering check, not fresh held-out evidence, and it is excluded
from the main ranking.

\subsection{Cost and latency}
\label{sec:cost}

Rules makes no LLM call. The LLM and hybrid variants make one call per
built workflow, skipping generation when clarification is requested.
Historical end-to-end latency is approximately $1.7$--$2.0$ seconds per
workflow. A new local microbenchmark times only the frozen enforcement
function, taking the median of 20 repetitions for each held-out
candidate. The median across candidates is $2.3$ microseconds on the
recorded Python environment. This excludes parsing, generation, and
execution. We do not infer billed API costs from output character counts.

\section{Analysis}
\label{sec:analysis}

\subsection{Static proxy and execution outcomes}

Across primary held-out runs, excluding enforcement-off and post-hoc
variants, 715 system--prompt pairs have a definite local outcome.
The static proxy passes 266 of these; none executes unsafely under the
percentage-point interpretation in evaluator v2. This is empirical
agreement within the checker, not a soundness proof. Another 345 pairs
fail the proxy but have a non-unsafe outcome, including protective
refusals and unfilled orders. The remaining 104 execute unsafely.

The stress test explains why these measurements must stay separate.
A candidate with a declared $50\%$ gate can pass a presence-based static
check and still execute above the fixed $5\%$ cap. Evaluator v2 catches
this failure even when a gate is declared. Full-DAG reachability and
semantic fidelity remain outside the execution test.

\subsection{Sensitivity to threshold units}

Historical outputs contain bare thresholds whose units are unspecified.
For the 53 such held-out primary outputs, interpreting numbers below one
as fractions instead of percentage points changes 17 outcomes; seven
execute unsafely under that alternative interpretation. Both analyses
are retained. This is evidence for an explicit unit contract at generation
time, not evidence that either convention recovers the model's intended
meaning in every case. The main tables consistently use percentage points.

\subsection{Mainnet arithmetic fidelity}
\label{sec:fidelity}

The execution test uses synthetic reserves. An earlier validation reads
Uniswap~V2 reserves at a pinned mainnet block and seeds the local AMM
with those values. Across 11 token pairs and seven trade sizes, from
$10^{-6}$ to $0.5$ of the input reserve, recorded quotes and local
executed outputs match the router's integer formula with zero relative
deviation. The per-pair grid is included in the artifact.

For a constant-product AMM with the same fee, impact depends on the
fraction of reserves traded. The measured impact is approximately $9.3\%$
at a tenth of the reserve and $33.5\%$ at half. Agreement on this
arithmetic does not validate full protocol routing, token behavior, or
adversarial transaction ordering. Mainnet interaction is read-only.

\subsection{Structural and configuration failures}

The offline Koan baseline misses required edges on 71 held-out prompts
and required nodes on 54. Direct Gemini omits the price-impact predicate
on 43 prompts, transaction monitoring on 32, and bridge confirmation on
15. These distinct error profiles explain why a single graph score
cannot characterize execution readiness.

Koan-Safe hybrid reduces missing safety predicates but retains structural
errors on novel workflows. Its graph coverage decreases from $1.00$
on easy held-out prompts to $0.77$ on medium and $0.25$ on hard prompts.
These strata use the benchmark's team-assigned difficulty labels.
The improvement over basic prompting does not remove the difficulty of
generating new graph structures.

\subsection{What the reference oracle establishes}

The oracle records 43 impact aborts and 17 slippage reverts among
60 definite held-out outcomes. Its generic amount of 100 and target
price of 3000 do not represent the intended trade for every prompt.
Its behavior therefore checks the scorer and execution path, but cannot
establish which user requests should result in a trade. Evaluator v2
records mined receipts for its submitted failures; we do not use this
synthetic oracle to claim a deployment-level gas-cost advantage.

\subsection{Implications}

The matched replay shows that structural repair and safety defaults
change both static coverage and local trade outcomes. The policy stress
test also shows why default injection is insufficient when an existing
parameter is unsafe. A workflow interface needs an explicit unit contract,
numeric-policy validation, and checks that protections actually control
the execution path. Our results measure the first two kinds of failure
under controlled configuration-level execution; full workflow validation
remains a separate requirement.

\section{Related work}
\label{sec:related}

\paragraph{Workflow generation and optimization.}
WorFBench~\cite{worfbench2025} evaluates agentic workflow generation
through graph structure and node/edge correctness.
Chat2Workflow~\cite{chat2workflow2026} targets executable visual
workflows, while FlowBench~\cite{flowbench2024} studies planning guided
by workflow representations. FlowMind~\cite{flowmind2023} generates
financial workflows using vetted APIs. We specialize workflow evaluation
to declared trade protections and controlled AMM outcomes.
Workflow optimization is a related but different task.
DSPy~\cite{dspy} optimizes modular language-model programs against a
metric, and AFlow~\cite{aflow} searches code-represented agentic
workflows using execution feedback. Koan-Safe instead constructs a
DeFi workflow for each request with a frozen generator and repair layer;
it does not search for an optimized agent program.

\paragraph{Tool agents and outcome-based evaluation.}
ReAct~\cite{react} interleaves reasoning and actions using observations
from an environment. ToolLLM~\cite{toolllm2024} and
API-Bank~\cite{apibank2023} evaluate API selection and use, while
LLMCompiler~\cite{llmcompiler} plans dependencies and schedules parallel
function calls. For evaluation, $\tau$-bench~\cite{taubench}
compares final database states with annotated goals under domain policies
and measures consistency over repeated trials. Our static proxy cannot
replace such an end-state test. We report one saved generation per
prompt, followed by controlled replay, rather than repeated-generation
reliability or success in a multi-turn user interaction.

\paragraph{Safety and security benchmarks for agents.}
ToolEmu~\cite{toolemu} evaluates accidental risks from tool use in
LM-emulated environments, including ambiguous instructions.
Agent-SafetyBench~\cite{agentsafetybench} covers broader behavioral
risks across simulated interactions. AgentDojo~\cite{agentdojo}
tests prompt injection through untrusted tool outputs and checks
environment states; AgentHarm~\cite{agentharm} studies explicitly
malicious user requests and multi-step misuse. Our waiver prompts test whether a direct
request can remove a required trade protection. They do not establish
resistance to indirect prompt injection or general agent misuse.
The local EVM gives reproducible numeric outcomes for a narrow set
of configurations, at the cost of the broader environmental coverage
available in these benchmarks.

\paragraph{Separating generation from enforcement.}
LLM-Modulo~\cite{llmmodulo} argues for combining language models with
external model-based verifiers, rather than relying on self-verification.
GuardAgent~\cite{guardagent} is a closer agent-safety precedent: it
translates guard requests into plans and executable checking code.
Constitutional Classifiers~\cite{classifiers} place learned input/output
safeguards around a model and evaluate their refusal and compute costs.
Koan-Safe uses fixed, domain-specific graph repairs and safety defaults,
not a learned classifier or generated guard program. We measure this
repair layer's effect with the generated candidate held fixed.

Runtime enforcement also has a formal foundation.
Schneider~\cite{schneider} characterizes policies enforceable by execution
monitoring, and shield synthesis~\cite{shield} constructs monitors that
correct unsafe outputs subject to a formal specification. Koan-Safe
does not provide comparable guarantees. Its frozen layer edits workflows
before execution and can retain permissive thresholds; the checker
interprets configurations without verifying that every transaction path
passes through a guard. Agentproof~\cite{agentproof} addresses that
graph-level distinction by checking structural properties and temporal
policies over extracted agent workflows. Koan-Safe repairs requested
structures but does not perform its all-path policy verification.

\paragraph{Natural language and DeFi transactions.}
Intent2Tx~\cite{intent2tx2026} is the closest transaction-generation
benchmark. It includes single-step and multi-step intents and evaluates
generated Ethereum transactions on forked state.
\textsc{Intent-Tx-18K}~\cite{intenttx18k2026} instead evaluates alignment
between an intent and a proposed transaction, including adversarial
violations. Multi-step evaluation alone does not distinguish our work.
We study typed workflow graphs, declared protections, and matched
repair ablations. Our controlled local execution has less protocol
coverage and weaker intent validation than forked-mainnet evaluation.

PACE~\cite{pace} addresses a complementary DeFi enforcement problem.
Its signed policy decisions bind approved intents and simulation reports
to execution bytes, with authorization enforced by a smart account.
Our generated safety declarations do not create such a binding.
PACE's transaction-level authorization and Koan-Safe's pre-execution
workflow repair operate at different points in the pipeline; we do not
claim the latter substitutes for the former.

\paragraph{Economic and protocol safety.}
Uniswap~V2~\cite{uniswapv2} provides the constant-product exchange model
used by our checker. The DeFi risk taxonomy~\cite{sokdefi2022} and
Flash Boys~2.0~\cite{flashboys2020} describe risks beyond a transaction's
syntax, including interactions between protocols and transaction ordering.
DeFiRanger~\cite{defiranger} recovers higher-level DeFi semantics from
transactions to detect price-manipulation attacks. Our own-trade impact
test is narrower than attack detection. It does not model an adversary,
oracle manipulation, sandwich attacks, or bridge failures, and a
slippage bound alone is not a substitute for evaluating those risks.

\paragraph{Benchmark quality and software testing.}
BetterBench~\cite{betterbench} examines benchmark quality across design,
evaluation, and reporting. Datasheets~\cite{datasheets} motivate explicit
documentation of dataset composition, collection, and intended uses.
These concerns matter for our small, team-authored corpus; a method
freeze does not make its annotations independent. Our paired
transformations apply established metamorphic-testing
principles~\cite{metamorphic}: related inputs impose relations on
outputs when a complete test oracle is unavailable. Satisfying those
relations provides additional evidence, not proof of semantic correctness.

\section{Limitations}
\label{sec:limitations}

\paragraph{Execution scope.} The local test program interprets
configuration and declared predicates. It does not traverse the
generated DAG, verify that a gate dominates every execution path, or
invoke Koan's production executors. Static graph coverage also does
not reject cycles or every malformed extra edge. The results support
configuration-level trade checks, not a full-workflow safety guarantee.
Cross-chain execution is unsupported; compositional cases execute only
their swap leg. Limit orders use a target-price condition on a static pool.

\paragraph{Units and evaluator dependence.} Historical generation
prompts did not specify output units. Evaluator v2 uses percentage points
for impact thresholds and separately tests the fractional interpretation
of bare values below one. Some outcomes and metamorphic conclusions
depend on that choice. Future generation runs should enforce an explicit
unit-bearing schema. Static safety checks require declared predicates
and concrete values but do not establish numeric-policy validity or
agreement with the user's intent.

\paragraph{Scale and annotation.} The development and held-out sets
contain 120 and 87 team-authored prompts. The same team developed the
method and authored the held-out set. Although the method was frozen
first, this is not an independently authored distribution shift.
An independent second-annotation pass has not been completed. The
artifact includes a protocol and subset for that work, but we report
no inter-annotator agreement result.

\paragraph{Model variability.} The main model outputs come from a single
temperature-zero generation run per configuration. The matched ablation
controls candidate variation by replaying saved responses; it does not
measure variability across fresh model calls. Prompt-level intervals
and exploratory paired tests should be interpreted accordingly.
The hybrid's small numerical advantage over the LLM variant is
inconclusive.

\paragraph{Method and post-hoc extension.} The frozen layer adds missing
defaults but can retain permissive safety values and unsupported intent
fields from a candidate. The policy-cap extension addresses the tested
threshold failure, but was developed after inspecting evaluation results.
Its diagnostic performance cannot be presented as independent held-out
generalization. Neither version proves semantic fidelity to every request.

\paragraph{Protocol fidelity.} All transactions execute against mock
tokens and a local constant-product AMM. Mainnet reserve comparison
checks arithmetic, not real routing, adversarial ordering, liquidity
changes, token quirks, or bridge behavior. No live funds are used.

\section{Ethics and Responsible Disclosure}
\label{sec:ethics}

This work is an offline evaluation: no transaction in this paper touches
a live network or real funds. The benchmark is intended to help identify
unsafe generated workflows before execution, not to encourage executing
generated DeFi workflows with real assets. The evaluated no-code system
and the language models are used as-is through their public interfaces;
we report their failures to motivate safer synthesis, not to single out
any implementation. Model outputs and derived metrics are released so
the results can be independently reproduced and audited.

\bibliographystyle{IEEEtran}
\bibliography{references}

\end{document}